\documentclass[10pt,twocolumn,letterpaper]{article}

\usepackage{cvpr}              

\usepackage{graphicx}
\usepackage{amsmath}
\usepackage{amssymb}
\usepackage{booktabs}

\usepackage[pagebackref,breaklinks,colorlinks]{hyperref}

\usepackage[capitalize]{cleveref}
\crefname{section}{Sec.}{Secs.}
\Crefname{section}{Section}{Sections}
\Crefname{table}{Table}{Tables}
\crefname{table}{Tab.}{Tabs.}

\def\cvprPaperID{*****} 
\def\confName{CVPR}
\def\confYear{2022}

\begin{document}

\title{Multi-model Ensemble Learning Method for Human Expression Recognition}

\author{Jun Yu\\
University of Science and Technology of China\\
Hefei, China\\
{\tt\small harryjun@ustc.edu.cn}
\and
Zhongpeng Cai\\
University of Science and Technology of China\\
Hefei, China\\
{\tt\small czp$\_$2402242823@mail.ustc.edu.cn}
\and
Peng He\\
University of Science and Technology of China\\
Hefei, China\\
{\tt\small hp0618@mail.ustc.edu.cn}
\and
Guochen Xie\\
University of Science and Technology of China\\
Hefei, China\\
{\tt\small xiegc@mail.ustc.edu.cn}
\and
Qiang Ling\\
University of Science and Technology of China\\
Hefei, China\\
{\tt\small qling@ustc.edu.cn}
}


\maketitle

\begin{abstract}
Analysis of human affect plays a vital role in human-computer interaction (HCI) systems. Due to the difficulty in capturing large amounts of real-life data, most of the current methods have mainly focused on controlled environments, which limit their application scenarios. To tackle this problem, we propose our solution based on the ensemble learning method. Specifically, we formulate the problem as a classification task, and then train several expression classification models with different types of backbones—ResNet, EfficientNet and InceptionNet. After that, the outputs of several models are fused via model ensemble method to predict the final results. Moreover, we introduce the multi-fold ensemble method to train and ensemble several models with the same architecture but different data distributions to enhance the performance of our solution. We conduct many experiments on the \textit{AffWild2} dataset of the ABAW2022 Challenge, and the results demonstrate the effectiveness of our solution.
\end{abstract}

\section{Introduction}
\label{sec:intro}

The analysis of human affect, which aims to transfer the understanding of human feelings to computers, has been a topic of major research in human-computer interaction (HCI) systems. Human affective states can be decided by a wide range of sources in three main categories: visual, auditory and biological signals. Among these three categories, visual information is the most important data in analyzing human affect. Ekman \cite{Prince} was the first to systematically study human facial expressions who introduced six basic emotions, i.e., anger, disgust, fear, happiness, sadness and surprise. Furthermore, facial expressions are related to specific movements of facial muscles, called Action Units (AUs). The Facial Action Coding System (FACS) was developed, in which facial changes are described in terms of AUs \cite{Darwin}.

The expression recognition task has been under exploration for a long time, and many effective methods are proposed with remarkable performance. However, most of them only concentrate on the controlled environments \cite{Kamachi, ZHANG, Zheng, Jiang} and few of them propose effective solution for the in-the wild scenarios. And in fact, there are few datasets collected in the wild environments. To deal with these problems, the ABAW2022 challenge \cite{kollias2021analysing, kollias2021distribution, kollias2021affect, kollias2019expression, kollias2019face, kollias2019deep, zafeiriou2017aff} is held to promote the development of the in-the-wild expression recognition task with a new video dataset \textit{AffWild2} \cite{Zafeiriou}. The dataset includes 2,603,921 frames from 548 different videos. Each frame is annoted with one of the 8 expressions(6 basic expressions, neutral and others). The subjects in the daaset are 431 in total(265 male and 166 female). All the videos are captured in the wild, and is thus the dataset is suitable for the corresponding research.

\begin{figure*}[!t]
	\centering
    \includegraphics[width=1\textwidth]{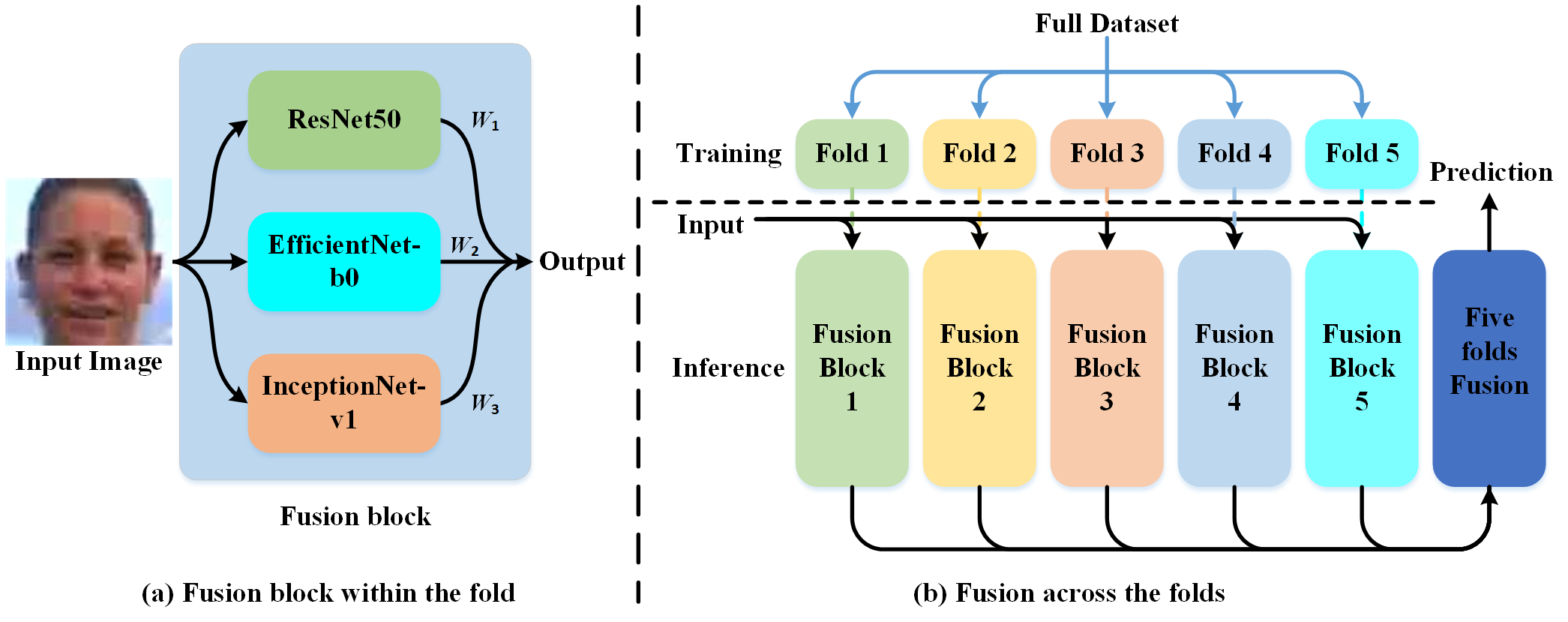}
    \caption{The overall pipeline of our method. It consists of two components: (a) Fusion block within the fold, (b) Fusion across the folds.}
    \label{overall}
\end{figure*}

In this work, we propose our solution based on the ensemble learning to deal with the problems of inconspicuous facial features and blurred expression classification boundaries in \textit{AffWild2} dataset. In order to fully extract facial features from each frames of the videos and improve the sensitivity of recognition to different emotional expressions, we formulate the problem as a classification task and train several human expression classification models with different types of backbones—ResNet \cite{Kaiming}, EfficientNet \cite{EfficientNet} and InceptionNet \cite{Szegedy} to get the outputs, which are fused via model ensemble method to predict the fused results. Considering the impact of different data distributions on model training, we introduce the multi-fold ensemble method to train and ensemble several models with the same architecture but different data distributions to enhance the performance of our solution.

In sum, the contributions of this work are two-folds:

\begin{itemize}
\item We ensemble three models with different architectures to improve the efficiency of our proposed model in expression recognition task. Meanwhile, focal loss \cite{Focal} is introduced to deal with the imbalanced problem in the \textit{AffWild2} dataset.
\item We split the whole training set into five folds and then train several models of the same architecture but with different folds. After that, we fuse the output of models trained with each fold via model ensemble method to further enhance the performance.
\end{itemize}

\section{Related Works}
Ensemble learning is an imperative study in machine learning \cite{Ensemble}. Over the previous years, ensemble learning has drawn considerable attention in the field of artificial intelligence, pattern recognition, machine learning, neural network and data mining. The ensemble method can be roughly divided into two categories, homogenous ensemble and heterogeneous ensemble. The homogenous ensemble means ensemble the models trained with the same algorithm, while the heterogeneous ensemble always ensemble models different in architectures or training methods. Huang et.al \cite{Huang} proposes the ensemble learning framework on face recognition tasks and concludes that ensemble method performs better than traditional method. Giacinto et.al \cite{Giacinto} introduced ensemble approaches for intrusion detection system and came to conclusion that the ensembled method has more excellent detection ability for known attacks than single model.

The benefits of the ensemble learning is that they are capable of boosting the weak learners so that overall accuracy of the learning algorithm on training data can be enhanced. Inspired by the fact, we constructs several human expressions classifiers and fuse their outputs so that the overall variance can be reduced and the evaluation result can be improved compared to the single classifier.

\section{Proposed Method}
In this section, we introduce our method for the 2nd track (Expression Classification Challenge) of the 3rd Workshop and Competition on Affective Behavior Analysis IN-THE-WILD(ABAW). The overall pipeline is illustrated in Figure \ref{overall}. The entire framework consists of two components: (a) Fusion block within the fold, (b) Fusion across the folds.

\subsection{Preprocessing}
In \textit{AffWild2} dataset, all videos are splitted into independent frames \cite{Dimitrios} and it provides the cropped and aligned images that are extracted from the videos. Each training image is resized to 112 $\times$ 112 and normalized before inputting into the models, and then we use the methods of RandomHorizontalFlipping and ColorJitter to enlarge our dataset. 

\begin{table*}[htb]   
	\begin{center}
	\setlength{\tabcolsep}{8mm}{   
	\caption{Five folds training results (F1-score) of three models.}  
	\label{table:1} 
		\begin{tabular}{c|c|c|c|c|c}   
		\hline   
		\textbf{Model} & \textbf{Fold 1} & \textbf{Fold 2} & \textbf{Fold 3} & \textbf{Fold 4} & \textbf{Fold 5}\\   
		\hline   
		ResNet50 & 0.317 & 0.232 & 0.227 & 0.266 & 0.271\\ 
		EfficientNet-b0 & 0.302 & 0.270 & 0.286 & 0.307 & 0.314\\
		InceptionNet-v1 & 0.275 & 0.242 & 0.236 & 0.297 & 0.300\\      
		\hline   
		\end{tabular}}
	\end{center}   
\end{table*}

\begin{table*}[htb]   
	\begin{center}
	\setlength{\tabcolsep}{1.2mm}{   
	\caption{Five folds training results (F1-score) of three model fusion methods.(InceptionNet-v1 : ResNet50 : EfficientNet-b0)}  
	\label{table:2} 
		\begin{tabular}{c|c|c|c|c|c}   
		\hline   
		\textbf{Fusion Method} & \textbf{Fold 1} & \textbf{Fold 2} & \textbf{Fold 3} & \textbf{Fold 4} & \textbf{Fold 5}\\   
		\hline   
		Fusion 1 & 0.331 (1:1:1) & 0.265 (1:1:1) & 0.246 (1:1:1) & 0.313 (1:1:1) & 0.326 (1:1:1)\\ 
		Fusion 2 & 0.323 (0.5:1.1:0.5) & 0.274 (0.6:0.4:1.2) & 0.280 (0.5:0.5:2) & 0.325 (0.6:0.4:1.1) & 0.331 (0.6:0.4:1.2)\\
		Fusion 3 & 0.324 (0.4:1:0.6) & 0.278 (0.6:0:0.7) & 0.285 (0.5:0:2) & 0.328 (0.4:0:0.6) & 0.326 (0.5:0:1.1)\\      
		\hline   
		\end{tabular}}
	\end{center}   
\end{table*}

\subsection{Model Structure}
As it shown in Figure \ref{overall}, the method we proposed consists of two components: (a) Fusion block within the fold, (b) Fusion across the folds.

For the fusion block within the fold module, We choose ResNet50 \cite{Kaiming}, EfficientNet-b0 \cite{EfficientNet} and InceptionNet-v1 \cite{Szegedy} as the backbones to extract the human facial expression features. In order to help the backbone learn more abundant facial features, we opt for the pretrained weights on the VGGFace2 dataset\cite{VGGFace} for ResNet50 and InceptionNet-v1, and use the pretrained weights on ImageNet\cite{ImageNet} for EfficientNet-b0, because the VGGFace2 dataset covers a large range of pose, age and ethnicity which can improve the robustness of our models. These three models have great differences in structure, so different features of the same image can be extracted, which is helpful for the classification of human expressions. By adjusting the fusion weights $W_{1}, W_{2}, W_{3}$ of the three networks, we can choose the best models to get the outputs, showing the advantages of ensemble learning.

As for the fusion block across the folds, as shown in the Figure \ref{overall}(b), we merge all the data in the training set and the validation set, and split the dataset according to the idea of five-fold cross-validation in the other module. This method can extract as much effective information as possible from existing data, obtain a more reasonable and accurate evaluation of the models, and avoid overfitting. Finally, we fuse the outputs of the five models to get the best performance of our method. 

\subsection{Loss Function}
We apply the Focal loss \cite{Focal} to deal with the imbalanced problem of the \textit{AffWild2} dataset. For the $m$ classes(include background), one-hot vector $y$ specifies the ground truth class $i (1\leq i\leq m)$ when $y_i=1$. The output $p$ of model for each voxel is a vector of length $m$, which representing the prediction probability for each class. Define the multi-class focal loss as:

\begin{equation} 
	\mathbf{F}\mathbf{L}(p)=\sum_{i=1}^{m}y_{i}\alpha_{i}\left(1-p_{i}\right)^{\gamma_{i}} log(p_{i})
\end{equation}

where super parameter $\alpha_i$ is the balanced weight of class $i$ and super parameter $\gamma$, which is called focusing parameter, is shared by each class.

\section{Experimental Results}

\begin{figure}[ht]
	\centering
    \includegraphics[width=0.45\textwidth]{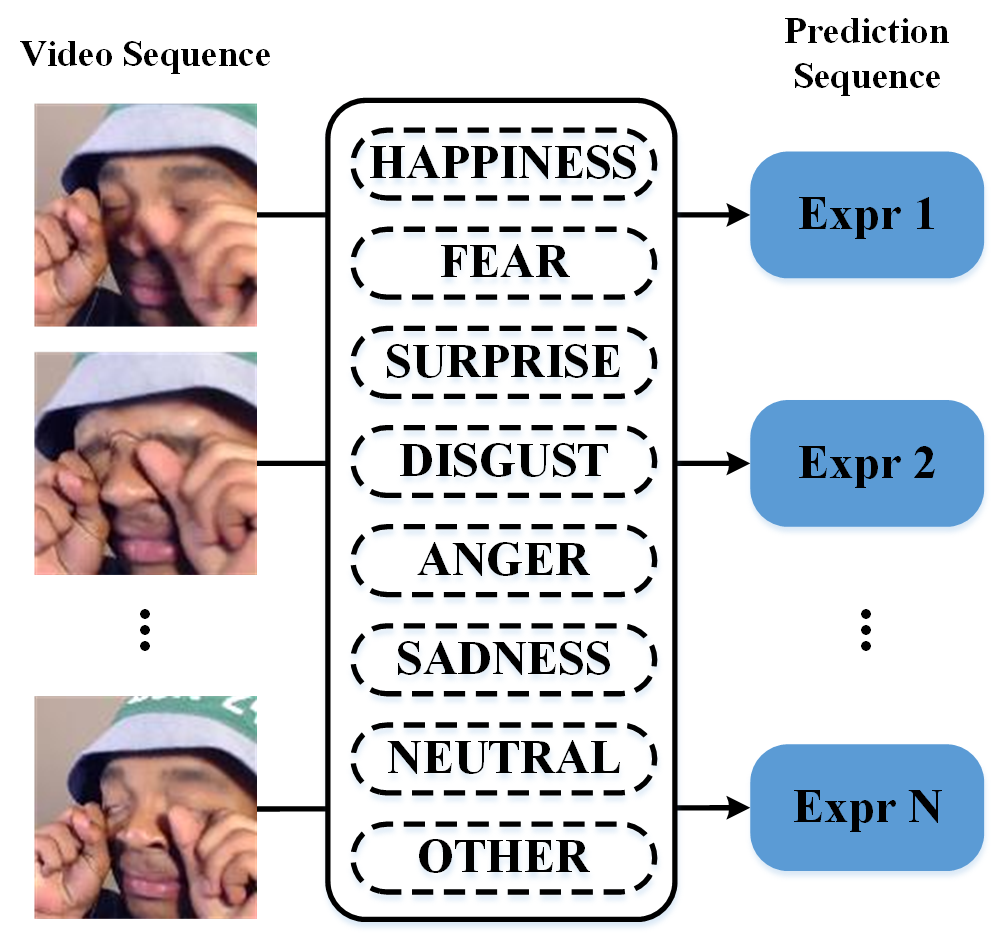}
    \caption{The process to get the expression prediction sequence.}
    \label{process}
\end{figure}

\subsection{Training Setup}
The training process is optimized by the Adam optimizer\cite{adam}. All the experiments are implemented on a NVIDIA Tesla A100 with a batch size of 256. To regularize the training process and accelerate the convergence of the model, we use the Cosine Annealing\cite{SGDR} as the learning rate scheduler with the initial learning rate of 0.001. Most of the testing models achieve the best validation index (F1-score) after 30 epochs training.

\subsection{Results}
We follow the process shown in Figure \ref{process} to get the prediction results for each frame in the video.

The performance index is the average F1-score across all 8 categories. The F1-score is a weighted average of the recall and precision and takes values in the range $[0, 1]$. It is defined as:

\begin{equation} 
	\mathbf{F_1}=\frac{2\times precision \times recall}{precision + recall}
\end{equation}

The evaluation criterion for the Expression Recognition Challenge is:

\begin{equation} 
	\mathbf{P}_{EXPR}=\frac{\sum_{expr}F_{1}^{expr}}{8}
\end{equation}

We merge the training set and the validation set and divide it into five subsets with similar size according to the data distribution characteristics. When conducting the five-fold experiment, four of the subsets are used as training set and the remaining subset is used as test set. By applying this division, we can conduct five experiments on the dataset and obtain five different trainning results.

Table \ref{table:1} shows the training results (F1-score) of three models——ResNet50, EfficientNet-b0 and InceptionNet-v1 in each fold experiment. Table \ref{table:2} shows the training results (F1-score) of different fusion methods. We conduct a series of experiments with different fusion weights and record the weights configuration of the best model. According to the final results, the indicators have been improved on the \textit{AffWild2} dataset and it can be noticed that the multi-model ensemble learning method plays an important role in the human expressions recognition task. Our final model will fuse the best results of each fusion method, which will further improve the final metric.

\section{Conclusion}
In this paper, we propose an ensemble learning method to deal with the chellange of human expression recognition by training several expression classification models with different types of backbones—ResNet, EfficientNet and InceptionNet and fuse the outputs of several models to predict the final results. Moreover, we introduce the multi-fold ensemble method to train and ensemble several models with the same architecture but different data distributions to enhance the performance. We conduct many experiments on the \textit{AffWild2} dataset of the ABAW2022 Challenge, and the results demonstrate the effectiveness of our solution.

{\small
\bibliographystyle{ieee_fullname}
\bibliography{egbib}
}

\end{document}